\documentclass[preprints,article,accept,moreauthors]{Definitions/mdpi}
\AtBeginDocument{\setlength{\fulllength}{\textwidth}}

\firstpage{1} 
\pubvolume{1}
\issuenum{1}
\articlenumber{0}
\pubyear{2026}
\copyrightyear{2026}
\externaleditor{Firstname Lastname} %
\datereceived{ } 
\daterevised{ } %
\dateaccepted{ } 
\datepublished{ } 

\Title{OBB-PBD: Extending Parallel Box Decoding to Oriented Bounding Boxes for Referring Expression Grounding in Aerial Imagery\texorpdfstring{$^{\dagger}$}{}}

\Author{Chang Woo Song %
  and Yeong %
 Jun Koh *}
\AuthorNames{Chang Woo Song and Yeong Jun Koh}
\address[1]{Department of Computer Science \& Engineering, Chungnam National University, Daejeon 34134%
, Republic of Korea; 202460116@o.cnu.ac.kr %
}
\firstnote{Published in \textit{Remote Sensing} \textbf{2026}, \textit{18}(20), 3535, \url{https://doi.org/10.3390/rs18203535}, under the Creative Commons Attribution (CC BY 4.0) license. This version has the same content as the published article; only the layout differs.}
\corres{\hangafter=1 \hangindent=1.05em \hspace{-0.82em} Correspondence: yjkoh@cnu.ac.kr}

\abstract{Referring expression grounding in overhead imagery requires oriented bounding boxes rather than horizontal ones,
since aerial objects appear at arbitrary orientations. Most grounding models emit box coordinates sequentially;
Parallel Box Decoding (PBD) instead predicts all geometry slots in one forward pass, but only for four-coordinate horizontal
boxes. We extend this block-aligned interface to a six-slot oriented representation using a double-angle
parameterization. Rather than assuming correctness from a decreasing loss curve, we verify the extension with
image/prompt-shuffle controls and visibility/leakage probes, which show that the corrected model depends on both
the image and the prompt rather than on prompt statistics alone. On the aerial vehicle visual grounding (AVVG) benchmark, whose
targets are genuinely oriented, the extended model improves rotated intersection over union (rotIoU) from 0.1253 for the zero-shot baseline to $0.303 \pm 0.012$ (mean $\pm$ standard deviation over three training seeds) across
the complete 6139%
-record test split spanning all 120 images. Learning oriented boxes rather than their axis-aligned envelopes adds $0.055 \pm 0.008$ rotIoU beyond a
matched control (per-seed 95\% confidence intervals all excluding zero), whereas block expansion alone yields no
rotIoU gain distinguishable from zero.}

\keyword{oriented bounding box; referring expression grounding; parallel box decoding; vision-language model; UAV imagery; double-angle parameterization; remote sensing visual grounding}

\addhighlights{yes}
\renewcommand{\addhighlights}{%
\\
\textbf{What are the main findings?}
\begin{itemize}[labelsep=2.5mm,topsep=3pt]
\item We extend Parallel Box Decoding, which predicts all bounding-box coordinates in a single forward pass, from four-coordinate horizontal boxes to a six-slot oriented representation. We further validate the extension using image/prompt-shuffle controls and visibility/leakage probes rather than relying solely on the training loss.
 \item On the aerial vehicle visual grounding (AVVG) benchmark, whose targets are genuinely oriented, the extended model improves rotated intersection over union (rotIoU) from 0.1253 for the zero-shot baseline to 0.303 $\pm$ 0.012 (mean $\pm$ standard deviation over three training seeds) across the complete 6139%
-record test split. Further analysis indicates that part of the improvement comes from learning oriented boxes instead of their axis-aligned envelopes (+0.055 $\pm$ 0.008 rotIoU, with each seed's 95\% confidence interval excluding zero), not merely from the larger coordinate representation.
\end{itemize}
\textbf{What are the implications of the main findings?}
\begin{itemize}[labelsep=2.5mm,topsep=3pt]
\item Diagnostic controls such as shuffle tests and visibility/leakage probes can reveal defects that benchmark accuracy alone may fail to detect, providing a verification framework for other block-parallel decoding extensions.
\item Most of the improvement comes from the joint effect of domain adaptation and a change of decoding path, but learning oriented geometry adds a consistent gain that the format change alone does not provide, supporting oriented bounding boxes as a more appropriate target than horizontal boxes for referring expression grounding in overhead and aerial imagery.
 \end{itemize}
}

\begin{document}

\section{Introduction}

Referring expression grounding---%
localizing the object referred to by a natural-language expression---has been
studied extensively on natural images \citep{kamath2021mdetr,li2022glip}, and~increasingly on remote-sensing and
unmanned aerial vehicle (UAV) imagery \citep{zhan2023rsvg,kuckreja2024geochat,zhou2024geoground}. The~remote-sensing setting is harder for
reasons largely absent from natural-image benchmarks: scenes are dense with many similar objects (rows of
vehicles, repeated structures), targets are small relative to the image, and~referring expressions often depend
on relational or comparative language (``the car closest to the camera'') that cannot be resolved from box
geometry alone. A~further difficulty specific to overhead imagery is that objects appear at arbitrary in-plane
orientations. A~horizontal bounding box (HBB), the~output format most grounding systems still use, fits an
elongated, rotated object poorly, as~the aerial-detection literature has documented at length
\citep{xia2018dota,xie2021oriented}. An~oriented bounding box (OBB), parameterized by center, size, and~orientation, is the natural alternative when the referred object has a meaningful aspect ratio and~heading.

Most vision-language grounding models emit box coordinates as autoregressively generated tokens, one coordinate
at a time. This costs one decoding step per coordinate and treats a geometrically atomic quantity---one box---as
an arbitrarily ordered token sequence. LocateAnything \citep{wang2026locateanything}
addresses this with Parallel Box Decoding (PBD): a block-aligned interface in which a trigger token is followed by a
single forward pass that predicts every coordinate slot together, rather than one token at a time. This is
faster and keeps the coordinate representation aligned with the geometric unit it describes, but~the released
formulation covers only four-coordinate horizontal boxes. Whether this block-aligned interface extends to a
six-slot oriented representation---and what computational and accuracy trade-offs that extension entails---was not examined by the original method and is the main focus of this study. Separately, joint-regression
architectures such as SATGround \citep{toker2025satground} predict OBB parameters with a dedicated regression head
rather than a coordinate-token interface. This is a different architectural design from the block-aligned
decoding extended here, and~the two are not compared head-to-head in this study (Section~\ref{sec:limitations}).

Extending the PBD interface from four coordinates to six is not merely a matter of changing a configuration
value. The~block-alignment guarantee of PBD---that coordinate slots see the image and prompt through an intact
causal prefix---can fail silently: in the pretrained backbone's training path, the~coordinate slots lost
visibility of the image and prompt while the training loss still decreased normally. Such a failure is invisible
in a loss curve and was found only with targeted diagnostic probes (Section~\ref{sec:upstream-defects}), which is
why this study treats verification of the extension as part of the method rather than as an~afterthought.

This paper makes three contributions:

\begin{enumerate}
  \item \textbf{A structurally validated extension of PBD to oriented boxes%
}: the six-slot geometry block is
        confirmed to align with the underlying architecture's coordinate interface, and~the codec's round-trip
        error is measured separately from model accuracy, so that representation error is distinguished from
        grounding error (Section~\ref{sec:parallel-obb-decoding}).
  \item \textbf{A diagnostic protocol for block-parallel extensions}: image/prompt-shuffle controls and
        visibility/leakage probes test whether the model uses the image and whether the reported metric reflects
        the trained decoding path (Section~\ref{sec:diagnostic-checks}); applied to this backbone, they expose and
        verify the correction of a silent loss of visual conditioning (Section~\ref{sec:upstream-defects}).
  \item \textbf{A measured, bounded performance result} under this validated protocol: the extended model improves
        over the zero-shot backbone on the complete test split of the aerial vehicle visual grounding (AVVG)
        benchmark, with~checkpoint selection kept disjoint from that split (Section~\ref{sec:evaluation-protocol}).
        A three-step decomposition separates the improvement into the joint effect of domain adaptation and a
        decoding-path change, the~six-slot
        block format alone, and~learning oriented geometry (Section~\ref{sec:hbb-control}); scope and limitations
        are discussed in Section~\ref{sec:limitations}.
\end{enumerate}

The remainder of this paper is organized as follows. Section~\ref{sec:related-work} reviews related work
across natural-image grounding, remote-sensing grounding, oriented aerial detection, and~structured coordinate
decoding. Section~\ref{sec:method} formulates the task and describes the OBB-PBD extension, training
configuration, and~evaluation protocol. Section~\ref{sec:results} reports the main quantitative results, the~HBB-envelope control that isolates the format-switch effect from the orientation-learning effect, the~image-dependence diagnostic, and~efficiency. Section~\ref{sec:discussion} interprets these results and
discusses limitations, and~Section~\ref{sec:conclusions} concludes.

\section{Related~Work}

\label{sec:related-work}
\unskip

\subsection{Visual Grounding on Natural~Images}
\label{sec:rw-natural-grounding}

Visual grounding---localizing the object referred to by a natural-language expression---has mostly been
studied as text-conditioned detection on natural images. The~task originates from crowd-sourced
referring-expression benchmarks such as ReferItGame \citep{kazemzadeh2014referitgame} and the RefCOCO,
RefCOCO+, and~RefCOCOg splits \citep{yu2016modeling,mao2016generation}, which pair a natural-language description
with a single target region in everyday photographs. The~detection transformer (DETR)
\citep{carion2020detr} later provided an end-to-end, set-prediction backbone that grounding-specific models build
on. Modulated DETR (MDETR) \citep{kamath2021mdetr} trains an end-to-end detector on aligned text-image pairs, jointly reasoning
over phrases and image regions to support both phrase grounding and referring expression comprehension. Grounded language-image pretraining (GLIP)
\citep{li2022glip} unifies object detection and phrase grounding as a single grounded pretraining objective at
scale, showing that a shared text-image formulation transfers across detection and grounding benchmarks. More
recent open-vocabulary detectors and multimodal large language models extend this lineage further: Grounding DINO
\citep{liu2023groundingdino} combines a DETR-style detector with grounded pretraining for open-set detection,
while Kosmos-2 \citep{peng2023kosmos2}, Shikra \citep{chen2023shikra}, and~Ferret \citep{you2023ferret} equip
general-purpose vision-language models with coordinate grounding by representing boxes or points as tokens or
continuous features the language model can emit or consume. All of the above establish the text-conditioned
detection lineage that remote-sensing grounding methods build on, but~none addresses oriented geometry or a
parallel, block-aligned coordinate~decoder.

\subsection{Remote-Sensing Visual~Grounding}
\label{sec:rw-rsvg}

Remote-sensing visual grounding (RSVG) adapts the grounding problem to aerial and satellite imagery, where small
targets, dense scenes, and~orientation variation make localization substantially harder than in natural images.
RSVG \citep{zhan2023rsvg} defines the task with image-expression-box triplets on DIOR-RSVG using horizontal
bounding boxes, starting the remote-sensing grounding benchmark lineage without oriented annotations. Later remote-sensing vision-language models add oriented or multi-format grounding: GeoChat
\citep{kuckreja2024geochat} is a region-level dialogue model whose remote-sensing setting includes oriented
annotations, and~GeoGround \citep{zhou2024geoground} unifies horizontal-box, oriented-box, and~mask grounding in
one model, introducing the RefGeo instruction-following collection this study's AVVG subset is drawn from.
GeoChat and GeoGround therefore already demonstrate that oriented-box remote-sensing grounding is~achievable.

This grounding capability belongs to a broader wave of remote-sensing vision-language models built by
instruction-tuning a general-purpose backbone on remote-sensing-specific data. RSGPT \citep{hu2023rsgpt} provides
image-captioning and visual question answering benchmarks and training data, and~VRSBench \citep{li2024vrsbench}
adds visual grounding to those two tasks. SkyEyeGPT \citep{zhan2024skyeyegpt}, EarthGPT \citep{zhang2024earthgpt}, and~LHRS-Bot
\citep{muhtar2024lhrsbot} extend instruction tuning to multi-task or multi-sensor remote-sensing understanding, in~which visual grounding is one of several supported tasks rather than the primary target. A~related but distinct
output formulation is referring segmentation: RRSIS \citep{yuan2023rrsis} predicts a pixel-wise mask rather than a
box. None of these systems is evaluated here under
this study's PBD coordinate-block protocol, and~their reported numbers use different datasets, splits, and~metrics, so they are not directly comparable to the AVVG results in Section~\ref{sec:main-results}.

\subsection{Oriented Object Detection in Aerial~Imagery}
\label{sec:rw-oriented-detection}

A parallel line of work treats oriented bounding boxes as a detection-geometry problem rather than a
language-grounding one. DOTA \citep{xia2018dota} established a large-scale aerial object detection benchmark with
quadrilateral annotations, motivating oriented representations by documenting the scale, orientation, and~aspect-ratio variation that horizontal boxes fail to capture in aerial scenes; FAIR1M \citep{sun2021fair1m} and
HRSC2016 \citep{liu2017hrsc2016} extend this benchmark lineage to fine-grained category labels and ship-specific
imagery, respectively, each also using oriented annotations. A~large family of detectors builds directly on these
benchmarks: the region-of-interest (RoI) Transformer \citep{ding2019roitransformer} learns a spatial transformation of region proposals
under OBB supervision, Gliding Vertex \citep{xu2020glidingvertex} regresses relative vertex offsets on a
horizontal box instead of an angle, S$^2$A-Net \citep{han2021s2anet} aligns convolutional features to refined
oriented anchors, ReDet \citep{han2021redet} incorporates rotation-equivariant backbone features, and~Oriented
R-CNN \citep{xie2021oriented} is a representative high-performance detector built directly around oriented
proposals and regression. A~separate line of work addresses the angle-regression discontinuity directly; the same
discontinuity motivates this study's double-angle parameterization (Section~\ref{sec:task-formulation}). SCRDet
\citep{yang2019scrdet} adds an IoU-smooth L1 loss, which uses the intersection over union (IoU) between boxes to reduce the loss
discontinuity near the periodic boundary,
circular smooth label \citep{yang2020csl} recasts angle regression as classification over discretized bins, and~the Kullback--Leibler divergence (KLD) \citep{yang2021kld} and Gaussian Wasserstein distance (GWD)
\citep{yang2021gwd} losses instead convert the rotated box into a Gaussian
distribution and regress a distribution distance that is well behaved at the periodic boundary. None of these
methods is a text-conditioned grounding system. They motivate the oriented-geometry problem that this study
addresses within a language-grounding model rather than a closed-set detector, and~their reported metrics (mean average precision, mAP, under~detection-style matching) are not directly comparable to the rotated-IoU grounding protocol used~here.

\subsection{Structured and Parallel Coordinate~Decoding}
\label{sec:rw-parallel-decoding}

Representing structured outputs as token sequences is a broader trend: Pix2Seq \citep{chen2022pix2seq}
casts object detection as language modeling over a serialized sequence of quantized box-and-class tokens, and~OFA
\citep{wang2022ofa} unifies captioning, grounding, and~other vision-language tasks under the same
sequence-to-sequence interface. Both, like the grounding models discussed above, decode one token at a time. Two
lines of prior work instead pursue non-autoregressive or partially parallel decoding: in machine translation,
non-autoregressive neural machine translation \citep{gu2018nat} produces all output tokens in a single forward
pass at some cost in output quality, and~in object detection, Deformable DETR \citep{zhu2021deformabledetr}
predicts a fixed set of object queries in parallel through cross-attention to deformable sampling points rather
than a serialized coordinate sequence. Most vision-language grounding models, however, still emit box coordinates
as a sequence of autoregressively generated tokens, one coordinate at a time. LocateAnything
\citep{wang2026locateanything} instead introduces Parallel Box Decoding (PBD), treating a box as a single
structured unit whose geometry slots are predicted together in one block-aligned forward pass rather than in four sequential token predictions. LocateAnything's released formulation covers four-coordinate horizontal boxes;
it does not report an oriented extension or the angle-prediction behavior that orientation
introduces into the same block-aligned interface. This is the method that the present work extends.
Separately, SATGround \citep{toker2025satground} interfaces a remote-sensing grounding module through specialized
localization control tokens on a fine-tuned vision-language model, jointly reasoning over language and spatial information to
regress oriented box geometry. Because~a joint OBB regression head of this kind already exists in the
remote-sensing grounding literature, this study's contribution is specifically the PBD coordinate-block mechanism,
not joint OBB regression as such. This study does not compare against SATGround head-to-head under a matched
protocol; see the scope note in Section~\ref{sec:limitations}.

\subsection{Positioning}
\label{sec:rw-positioning}

This study investigates the oriented-box extension of the PBD interface introduced by LocateAnything
\citep{wang2026locateanything}, adding center, size, and~double-angle orientation slots to the same
block-aligned interface and evaluating the result under a validated implementation and evaluation protocol
(Sections~\ref{sec:parallel-obb-decoding} and~\ref{sec:evaluation-protocol}). Building on the oriented-box grounding capability GeoChat and GeoGround
already establish (Section~\ref{sec:rw-rsvg}) and the structured decoding LocateAnything already provides
(Section~\ref{sec:rw-parallel-decoding}), the~contribution here is an empirically validated architectural
extension together with a diagnostic evaluation protocol: integrating orientation into a LocateAnything-style,
block-aligned parallel decoding interface and characterizing its accuracy and latency trade-offs under a controlled
protocol (Sections~\ref{sec:main-results}--\ref{sec:efficiency}).

\section{Materials and~Methods %
}

\label{sec:method}
\unskip

\subsection{Task~Formulation}
\label{sec:task-formulation}

We study referring expression grounding in overhead UAV imagery. Given an image and a natural-language description,
the model predicts one oriented bounding box (OBB) for the referred object. The~target geometry is represented by
its center, width, height, and~orientation. We use the double-angle parameterization
$(c_x, c_y, w, h, \sin(2\theta), \cos(2\theta))$, which avoids the discontinuity at the $0^{\circ}$/$180^{\circ}$
boundary while preserving the $180^{\circ}$ periodicity of a~rectangle.

Because a rectangle satisfies $(w,h,\theta) \equiv (h,w,\theta+90^{\circ})$, an~oriented bounding box admits two
equivalent $(w,h,\theta)$ parameterizations for the same physical box. We resolve this before encoding
by canonicalizing every box to $w \geq h$: %
\begin{equation}
\text{if } w < h: \quad (w,h) \leftarrow (h,w), \qquad \theta \leftarrow (\theta + 90^{\circ}) \bmod 180^{\circ}.
\label{eq:canon}
\end{equation}
Every %
 coordinate is quantized with the same uniform codec: $[0,1]$ is divided into 1000 equal-width intervals,
and their 1001 endpoints form the quantization levels, each represented by one coordinate token
(\texttt{<0>%
}, \dots, \texttt{<1000>}). A~value $v$ already normalized to $[0,1]$ (i.e.\
$c_x/W$, $c_y/H$, $w/W$, or~$h/H$) is encoded as the nearest of these 1001 levels and decoded by inverting that
scale:
\begin{equation}
q(v) = \mathrm{round}(v \cdot 1000) \in \{0,\dots,1000\}, \qquad v = q/1000.
\label{eq:codec-unit}
\end{equation}
A signed value $u \in [-1,1]$ (i.e.\ $\sin(2\theta)$ or $\cos(2\theta)$) is first rescaled to $[0,1]$ before applying
the same unit codec, and~inverted the same way in reverse:
\begin{equation}
q(u) = \mathrm{round}\!\left(\tfrac{u+1}{2} \cdot 1000\right), \qquad u = 2 \cdot (q/1000) - 1.
\label{eq:codec-signed}
\end{equation}
At decode time, the two orientation levels are read back to $\hat{s} = \sin(2\theta)$, $\hat{c} = \cos(2\theta)$ via
Equation~\eqref{eq:codec-signed}, and~the angle is recovered as
\begin{equation}
\theta = \left(\tfrac{1}{2}\,\mathrm{atan2}(\hat{s}, \hat{c})\right) \bmod 180^{\circ},
\label{eq:decode-theta}
\end{equation}
followed by the same $w \geq h$ canonicalization of Equation~\eqref{eq:canon} applied to the decoded $(w,h,\theta)$,
so a decoded box is always compared against the ground truth in the same canonical form it was encoded~in.

\subsection{Parallel OBB~Decoding}
\label{sec:parallel-obb-decoding}

\begin{figure}[H]
\centering
\includegraphics[width=\textwidth]{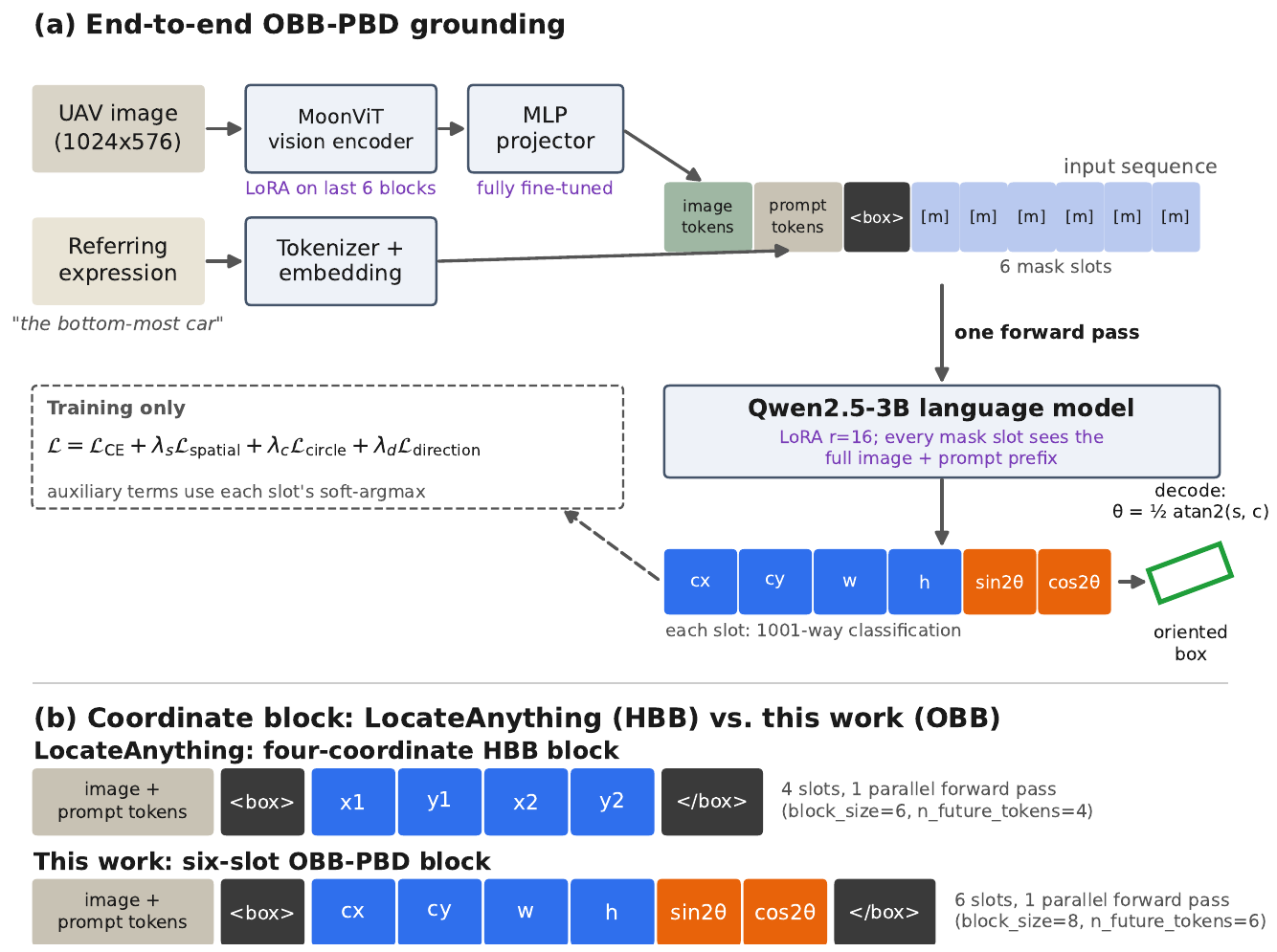}
\caption{OBB%
-PBD architecture. (\textbf{a}) End-to-end inference: MoonViT image features, projected by the multilayer perceptron (MLP)
projector, and~the tokenized referring expression form the prefix; the \texttt{<box>} trigger is followed by six
mask slots, and~a single forward pass of the language model predicts all six geometry tokens, which the codec of
Section~\ref{sec:task-formulation} decodes into an oriented box. Purple labels mark the adapted components
(Section~\ref{sec:training-configuration}); the dashed box shows the training objective
(Section~\ref{sec:training-configuration}). (\textbf{b}) The six-slot OBB-PBD block extends LocateAnything's
four-slot HBB block \citep{wang2026locateanything}: the first four slots switch from corner coordinates to
center-and-size, and~the two orange slots ($\sin(2\theta)$, $\cos(2\theta)$) are the only wholly new addition
(Section~\ref{sec:parallel-obb-decoding}). Solid arrows show the inference data flow; the dashed arrow indicates that the predicted slots enter the training objective.}
\label{fig:architecture}
\end{figure}

Our implementation extends LocateAnything's \citep{wang2026locateanything} four-coordinate horizontal-box (HBB)
block, which emits corner coordinates $(x_1,y_1,x_2,y_2)$, to~six parallel geometry slots parameterized by center,
size, and~orientation. Figure~\ref{fig:architecture} shows the resulting extension: the first four slots change
representation (corner $\rightarrow$ center-and-size), and~two orientation slots are added. Center-and-size
replaces corner coordinates because the double-angle orientation term is defined about the box center.
The emitted structure~is

\begin{quote}
\texttt{<box> $c_x$ $c_y$ $w$ $h$ $\sin(2\theta)$ $\cos(2\theta)$ </box>}
\end{quote}

The resulting configuration has \texttt{block\_size=8} and \texttt{n\_future\_tokens=6}: the opening \texttt{<box>}
token triggers one parallel coordinate forward pass, the~six geometry slots are predicted together, and~the closing
\texttt{</box>} token is generated at the following step. Each coordinate is represented with the existing
1001-level quantization codec of Equation~\eqref{eq:codec-unit}. The~codec's round-trip error is evaluated independently of model prediction: encoding and
decoding the ground-truth polygons directly yields a mean round-trip IoU of 0.9903 on the training split and 0.9823 on
the validation split (minimum 0.8379), indicating that quantization error is negligible relative to model
grounding~error.

\subsection{Backbone Corrections Required by the~Extension}
\label{sec:upstream-defects}

Because a block-alignment failure can corrupt training without appearing in the loss curve, the~six-slot
extension was validated with the diagnostic probes of Section~\ref{sec:diagnostic-checks} before any performance
metric was reported. One finding matters for the method: in the backbone's training path, the~causal-prefix
length that should expose the image and prompt tokens to the coordinate slots was computed from a position-ID
pattern that training never produces, so the coordinate slots saw none of the image or prompt tokens while the
loss still decreased. We recover the prefix length from the mask-token region, as~the inference path already does,
and the visibility probe confirms the fix: on the same training batch, the~first coordinate slot's attention
mask goes from 7 of 894 keys, none of them image or prompt tokens, to~all 894.
Four further defects in the backbone code caused outright failures (crashes) rather than silent ones; they are
also corrected and are listed in Appendix~\ref{app:defects}, and~the source changes are available as described in
the Data Availability~Statement.

The extended model itself is evaluated exclusively through direct decoding, and~the backbone's upstream
\texttt{generate()} path is used only to reproduce the zero-shot baseline as it would ordinarily be run; the
decoding-speed comparison in Section~\ref{sec:efficiency} uses its own cached autoregressive~implementation.

\subsection{Training~Configuration}
\label{sec:training-configuration}

We fine-tune the LocateAnything-3B backbone \citep{wang2026locateanything}---a Qwen2.5-3B-Instruct language model
\citep{yang2024qwen25} paired with a MoonViT vision encoder---on AVVG, the~RefGeo subset introduced by \citep{zhou2024geoground}. The~language model
uses low-rank adaptation (LoRA) \citep{hu2022lora} with rank 16, the~last six MoonViT vision blocks are adapted with LoRA, and~the multimodal
projector \texttt{mlp1} is
trained as a full module, so the vision pipeline is updated through both the MoonViT LoRA adapters and the
projector. The~projector's learning rate is one tenth of the adapters' ($1\times10^{-5}$), and~gradients are clipped
at norm 1.0. The~main training settings are given in Table~\ref{tab:training-settings}. Training used Python 3.11.10, PyTorch 2.5.1, Transformers 4.57.1, and PEFT 0.20.0 on three NVIDIA RTX 6000 Ada graphics processing units (GPUs; 48\,GB each; NVIDIA Corporation, Santa Clara, CA, USA).

The base loss $\mathcal{L}_{\text{CE}}$ is the backbone's standard next-token cross-entropy over the full response,
in which each of the six coordinate slots is itself a classification target over the 1001-way coordinate
vocabulary---the block format alone already turns coordinate prediction into token classification. Three
additional, differentiable terms are added on top of this discrete classification signal, each computed from a
soft-argmax expectation over a slot's level probabilities rather than from the hard argmax, so that all three remain
differentiable with respect to the logits. For~a slot with softmax probabilities $p(k)$ over the 1001 levels
$k \in \{0,\dots,1000\}$, the~soft-argmax estimate is $\hat{v} = \sum_k p(k)\, k / 1000$ (or, for~a signed slot,
the same sum rescaled by Equation~\eqref{eq:codec-signed}).

\begin{table}[H]
\caption{Main %
 training settings of the OBB-PBD~models.}
\label{tab:training-settings}

\begin{adjustwidth}{-\extralength}{0cm}

\setlength{\cellWidtha}{\fulllength/2-2\tabcolsep-2.20in}
\setlength{\cellWidthb}{\fulllength/2-2\tabcolsep--2.20in}
\scalebox{1}[1]{\begin{tabularx}{ \fulllength}{>{\raggedright\arraybackslash}m{\cellWidtha}>{\centering\arraybackslash}m{\cellWidthb}}
\toprule
\textbf{Setting} & \textbf{Value} \\
\midrule
Dataset                & AVVG subset of the RefGeo-derived data \citep{zhou2024geoground} \\
Training samples       & 26{,}451 usable records after removing 31 out-of-frame annotations, split into a
                          421-image training set (22{,}495 records) and a 74-image held-out validation set (3956
                          records) disjoint from the test split (Section~\ref{sec:evaluation-protocol}) \\
Image source           & 495 DJI UAV aerial images (421 training + 74 validation, disjoint from the 120-image
                          test split), $4000 \times 2250$ pixels \\
Optimizer              & AdamW (weight decay 0.01), constant learning rate $1\times10^{-4}$ (no schedule, no warmup) \\
Batch and precision    & 1 per GPU, gradient accumulation 4, 3 GPUs (global batch size 12);
                          bfloat16 (auxiliary losses computed in float32) \\
Loss weights           & $\lambda_{\text{spatial}}=2.5$, $\lambda_{\text{circle}}=0.25$, $\lambda_{\text{direction}}=0.5$ \\
Vision adaptation      & last 6 MoonViT blocks + full \texttt{mlp1} projector \\
Training budget        & 5 epochs; checkpoint saved after every~epoch \\
Run count               & three training seeds per model (seed sets initialization and data order; split fixed) \\
\bottomrule
\end{tabularx}}
\end{adjustwidth}
\end{table}

The \emph{spatial loss%
} applies a Smooth-L1 penalty between this soft-argmax estimate and the ground-truth target,
independently for each of the four spatial slots $c_x, c_y, w, h$ and averaged over them:
\begin{equation}
\mathcal{L}_{\text{spatial}} = \frac{1}{4} \sum_{i \in \{c_x,c_y,w,h\}} \mathrm{SmoothL1}_{\beta=0.01}(\hat{v}_i, v_i).
\label{eq:loss-spatial}
\end{equation}
The \emph{circular consistency loss} penalizes the predicted double-angle pair $(\hat{s},\hat{c})$ for departing
from the unit circle that a valid $(\sin(2\theta),\cos(2\theta))$ pair must lie on:
\begin{equation}
\mathcal{L}_{\text{circle}} = \left(\hat{s}^2 + \hat{c}^2 - 1\right)^2.
\label{eq:loss-circle}
\end{equation}
The \emph{directional loss} is a cosine distance between the predicted and ground-truth double- angle vectors,
\begin{equation}
\mathcal{L}_{\text{direction}} = \mathbb{1}[r_{\text{norm}} \geq 1.05] \left(1 - \frac{\hat{s}s + \hat{c}c}{\sqrt{\hat{s}^2+\hat{c}^2+\epsilon}\,\sqrt{s^2+c^2+\epsilon}}\right),
\label{eq:loss-direction}
\end{equation}
averaged over the samples with $r_{\text{norm}} \geq 1.05$ in the batch, where $(s,c)$ is the ground-truth pair,
$\epsilon=10^{-7}$, and~$r_{\text{norm}} = \max(w/W, h/H)/\min(w/W, h/H)$ is the image-normalized side ratio: the
ratio of the box's width and height after each is divided by the corresponding image dimension. It is not the
box's physical aspect ratio $w/h$; it equals 1 for a box whose shape matches the image, not for a square box. The~gate was designed to zero the directional gradient for near-square boxes, whose orientation is nearly undefined
and whose gradient is therefore close to noise. Because~AVVG's images are non-square ($4000\times2250$), the~gate
as defined instead excludes boxes whose aspect ratio is close to the image's (about 16:9), and~physically
near-square boxes still receive the directional loss. To~test whether this matters, we retrained OBB-PBD with the gate computed from the physical aspect ratio $w/h$, keeping seed 1's
initialization and data order so that only the gate changes. It reached a rotated IoU (rotIoU) of 0.3008 (95\% confidence interval (CI) $[0.2799, 0.3228]$),
lower than the 0.3124 of the otherwise identical seed-1 run with the normalized gate (paired difference $-0.012$,
per-record image-cluster CI $[-0.019, -0.004]$). That interval reflects test-set sampling only; with a single
physical-gate run, it cannot separate the gate's effect from training randomness, and~the result lies within the
spread of the three normalized-gate seeds (0.290--0.312). Correcting the gate therefore does not improve the
model, and~the normalized gate is retained.
The total training loss is
\begin{equation}
\mathcal{L} = \mathcal{L}_{\text{CE}} + \lambda_{\text{spatial}}\mathcal{L}_{\text{spatial}} + \lambda_{\text{circle}}\mathcal{L}_{\text{circle}} + \lambda_{\text{direction}}\mathcal{L}_{\text{direction}},
\label{eq:loss-total}
\end{equation}
with the $\lambda$ values of Table~\ref{tab:training-settings}.

RefGeo \citep{zhou2024geoground} consolidates five subsets (AVVG, RSVG, DIOR-RSVG, GeoChat, VRSBench), each with
its own image pool and train/test split files. This study uses only the AVVG subset, which includes oriented
bounding box annotations and provides complete native aerial imagery.
A data audit found that 97.85\% of the boxes are not axis-aligned and that the image file could be resolved for
100\% of the audited records; paired image and annotation examples were also rendered for visual verification
before training. Every training and evaluation script in this study therefore uses the AVVG subset, and~the other
four RefGeo subsets are not used anywhere in the reported~results.

Generative artificial intelligence (AI) tools assisted with this study's execution: Claude Code %
 (Anthropic; Claude Sonnet 5 and Claude Opus 5.5) and Codex 
 (OpenAI; GPT-6-Sol) assisted
with implementation, execution of author-approved experiments, and~manuscript drafting, and~Codex and Antigravity (Google; Gemini 3.8 Flash) were used for independent manuscript cross-checks. All research questions, methodological
choices, and~result interpretations remain the authors' own; see Acknowledgments for full~disclosure.

\subsection{Evaluation Protocol and Checkpoint~Selection}
\label{sec:evaluation-protocol}

We evaluate with the direct PBD decoder, which constructs the prompt, the~\texttt{<box>} token, and~six coordinate
mask slots and obtains the geometry logits in a single forward pass. This path matches the training block
definition. The~upstream \texttt{generate()} path implements multi-token prediction only for the original
six-token HBB block, so the six-slot model is evaluated through the direct~decoder.

The primary metric is rotated IoU (rotIoU): the IoU between the predicted and ground-truth oriented boxes,
computed from the polygon intersection of the two oriented quadrilaterals using the Shapely library (version 2.1.2) rather than
a custom clipping routine. We also report axis-aligned IoU (aabbIoU), the~IoU between the HBB envelopes (defined
below) of the predicted and ground-truth boxes, which compares with the original HBB baseline on its own terms.
The remaining metrics are threshold accuracies at
rotIoU 0.50 and 0.75 (Acc@0.50, Acc@0.75), angle mean absolute error (MAE) in degrees, and~the decoded percentage.
A decoding failure is assigned zero IoU and is included in the~denominator.

Because $\theta$ is only defined modulo $180^{\circ}$ after the canonicalization of Equation~\eqref{eq:canon}, a~naive $|\hat{\theta}-\theta|$ difference would overstate the error for pairs that straddle the $0^{\circ}$/$180^{\circ}$
wrap-around (e.g., \ $\hat{\theta}=1^{\circ}$ against $\theta=179^{\circ}$ is a $2^{\circ}$ error, not $178^{\circ}$).
Angle MAE is therefore the mean of the periodic distance
\begin{equation}
d(\hat{\theta},\theta) = \min\!\left(|\hat{\theta}-\theta| \bmod 180^{\circ},\ 180^{\circ} - \left(|\hat{\theta}-\theta| \bmod 180^{\circ}\right)\right)
\label{eq:angle-mae}
\end{equation}
over all samples with a successful decode. For~every trained run, the~reported checkpoint is the epoch with the
highest rotIoU on the held-out validation split defined in Table~\ref{tab:training-settings}. Each model is
trained with three seeds; a seed sets both parameter initialization and the data order, while the image-level
split is identical across seeds. For~OBB-PBD, validation rotIoU rose over the five epochs in seeds 1 and 2 and
plateaued between epochs 4 and 5 in seed 3 (0.2780 at both), so epochs 5, 5 and 4 are~selected.

The baseline is the zero-shot LocateAnything-3B backbone evaluated with its upstream \texttt{generate()} path on
the complete AVVG test split, and~serves as the primary HBB baseline. We define the \emph{HBB envelope} of a box
as the tightest axis-aligned rectangle enclosing it. A~reference derived from the ground truth (GT), the~\emph{GT-derived HBB envelope (reference)}---the HBB envelope of each target's GT oriented box---is reported
alongside the baseline; it is a computed reference, not a learned model. This rectangle is a natural, easily
computed reference point that illustrates the gap between horizontal and oriented representations, but~it is not
guaranteed to be the rotIoU-\emph{maximizing} axis-aligned rectangle for each target, so its score should be read
as a reference value rather than as a formal upper bound on what any axis-aligned prediction could achieve
(Section~\ref{sec:main-results}).

\subsection{Diagnostic~Checks}
\label{sec:diagnostic-checks}

We use three complementary checks beyond the main score:

\begin{enumerate}
  \item \textbf{Image dependence:} evaluate each selected checkpoint with matched image/prompt pairs,
        independently shuffled images, and~independently shuffled prompts (Section~\ref{sec:image-dependence}). A~        large matched-versus-image-shuffled gap is required before treating the score as evidence that the
        coordinate slots use visual input. This check is motivated by the broader concern that a high grounding
        score need not demonstrate use of every required input modality: \citet{cirik2018visual} show that a
        system given only the image, with~the referring expression withheld, still reaches 71.2\% top-2 precision
        on a standard benchmark, indicating that the benchmark's own label distribution---not language
        understanding---accounts for much of that score. Their ablation tests dependence on language; the image-shuffled
        condition here tests the complementary case, dependence on vision.
  \item \textbf{Implementation visibility and leakage probes:} verify that coordinate slots can attend to the
        image and that ground-truth coordinate tokens are not injected into the model input during training.
  \item \textbf{Qualitative overlays:} render predictions and target OBBs on the original UAV frames to
        distinguish geometric errors from referring failures, such as placing a box on a neighboring vehicle
        (Figure~\ref{fig:qualitative}).
\end{enumerate}

{\begin{figure}[H]
\centering
\begin{minipage}{0.315\textwidth}
\centering
\includegraphics[width=\textwidth]{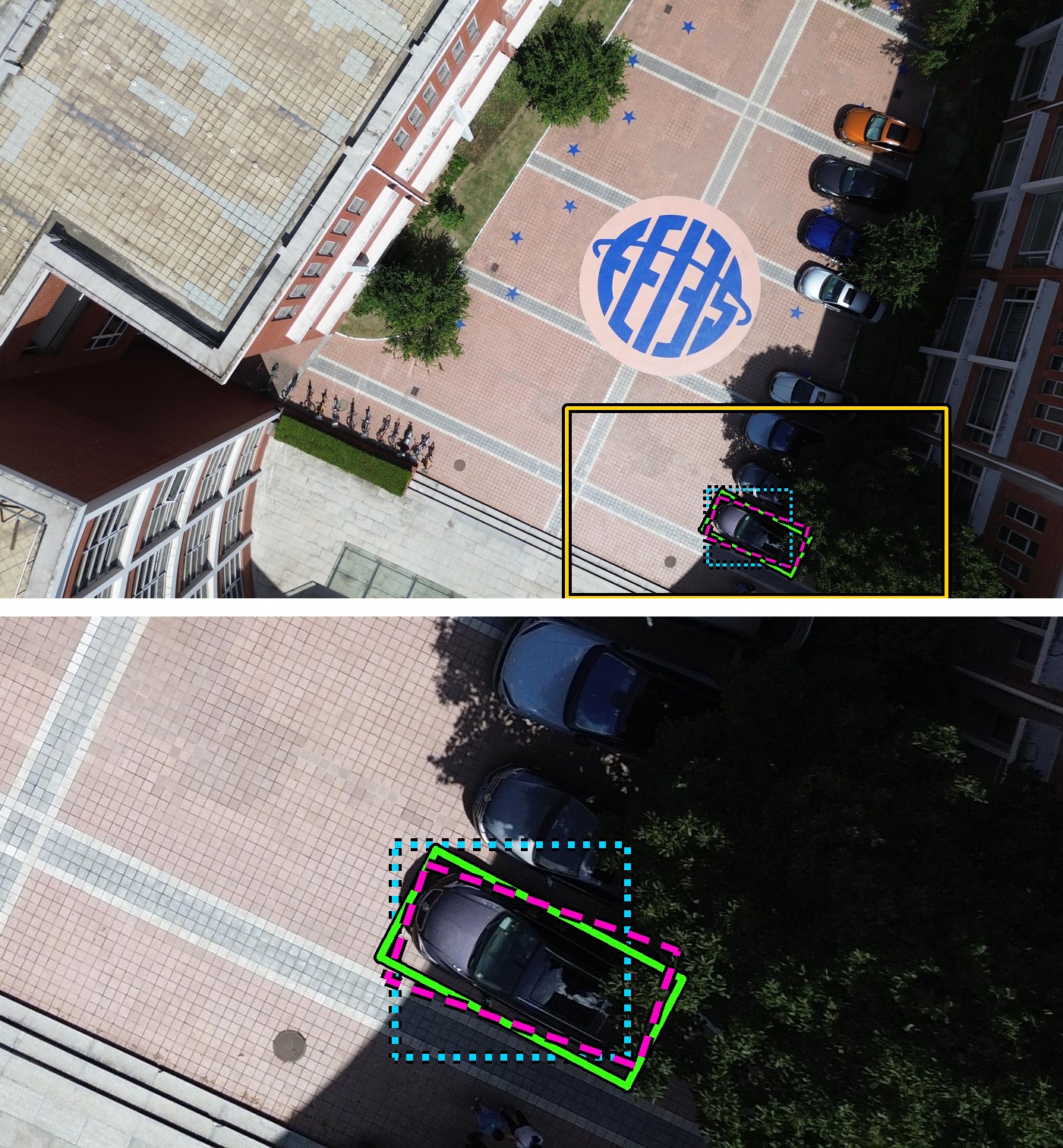}
\small (a) rotIoU 0.82
\end{minipage}
\hfill
\begin{minipage}{0.315\textwidth}
\centering
\includegraphics[width=\textwidth]{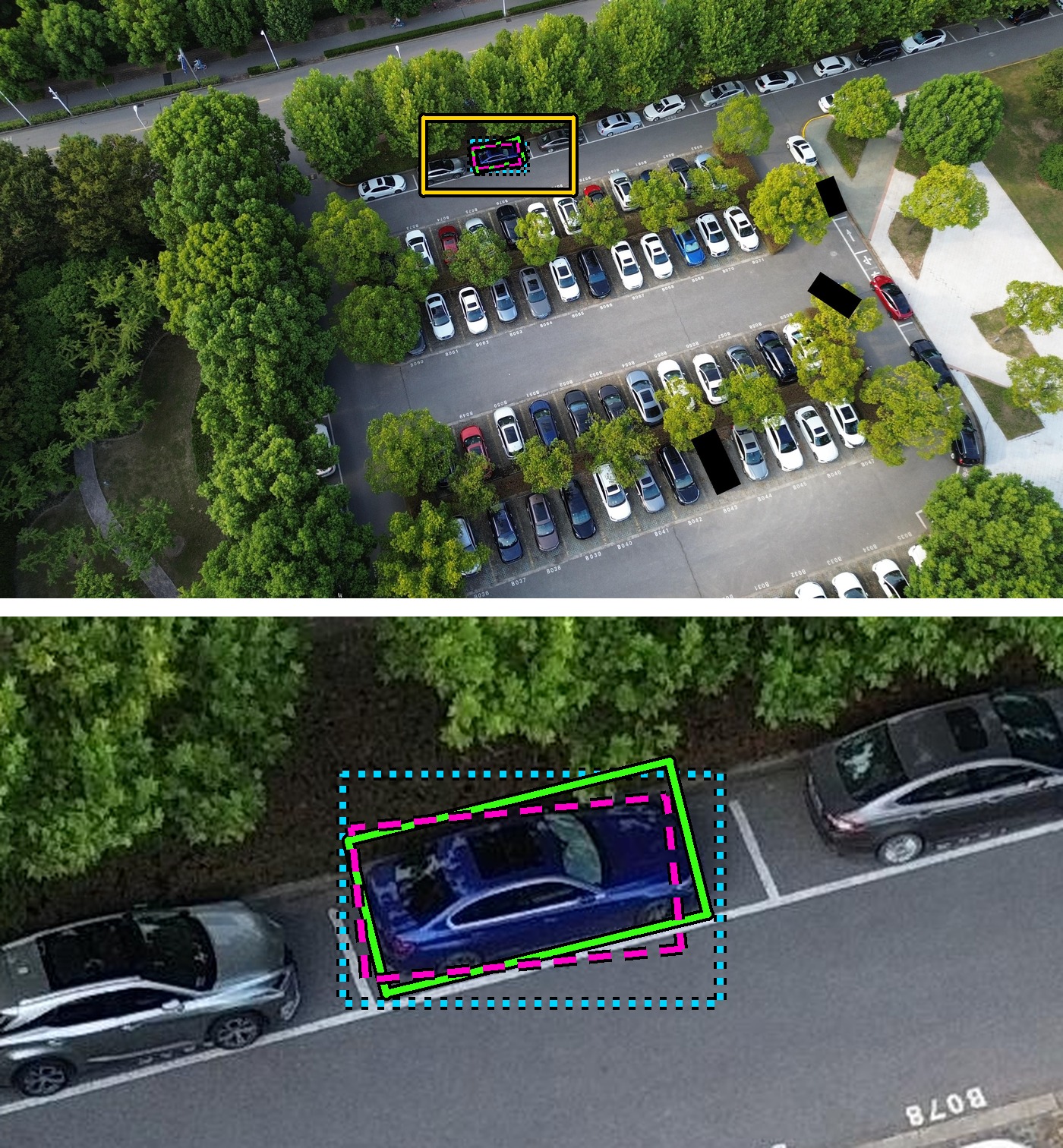}
\small (b) rotIoU 0.79
\end{minipage}
\hfill
\begin{minipage}{0.315\textwidth}
\centering
\includegraphics[width=\textwidth]{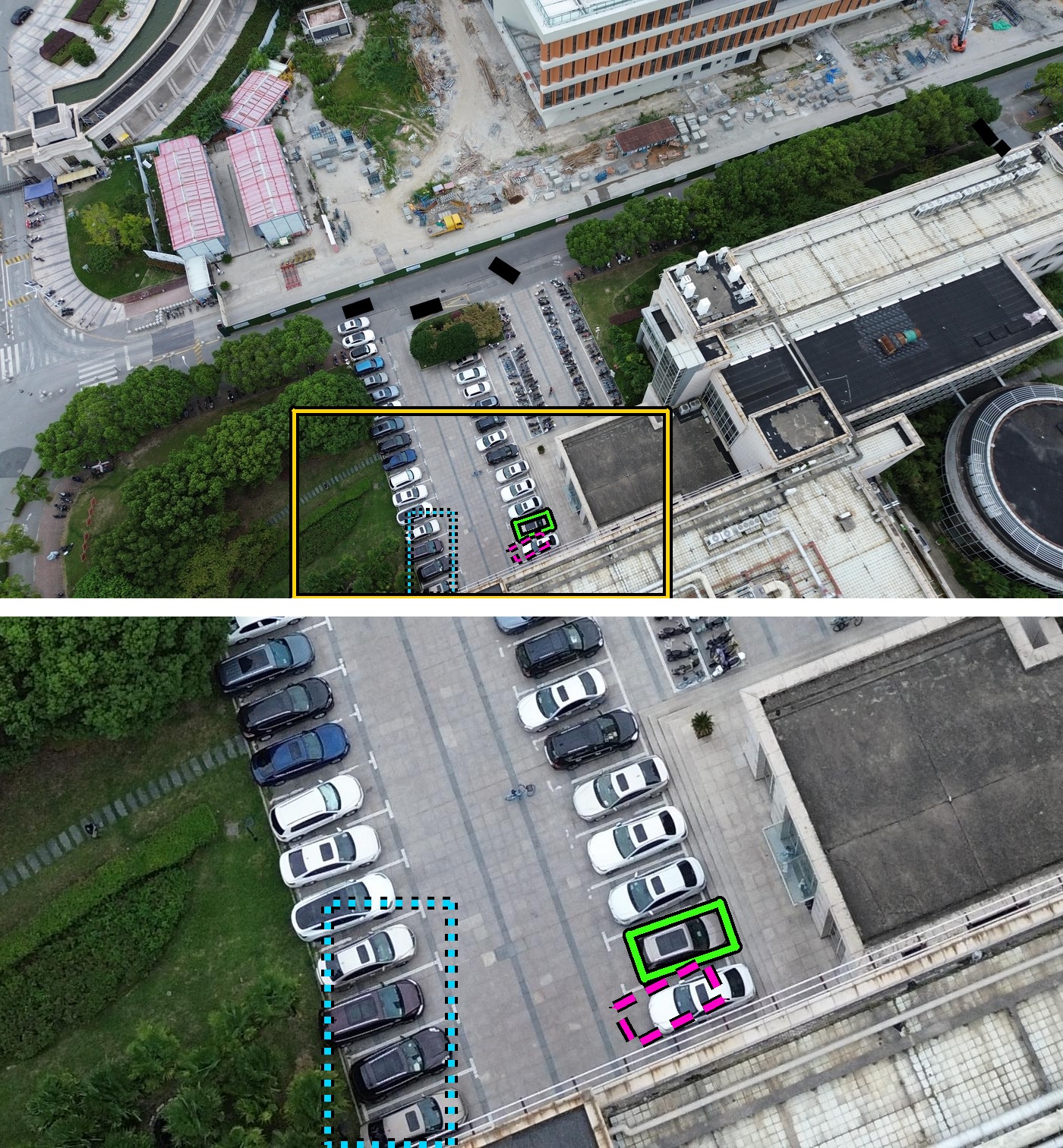}
\small (c) rotIoU 0.00
\end{minipage}

\vspace{4pt}
\footnotesize
\begin{tabularx}{\textwidth}{l*{10}{>{\centering\arraybackslash}X}}
\toprule
 & \multicolumn{6}{c}{\textbf{OBB-PBD}} & \multicolumn{4}{c}{\textbf{Baseline (HBB)}} \\
\cmidrule(lr){2-7}\cmidrule(lr){8-11}
\textbf{Panel} & $\Delta c_x$ & $\Delta c_y$ & $\Delta w$ & $\Delta h$ & $|\Delta\theta|$ ($^{\circ}$) & rotIoU & $\Delta c_x$ & $\Delta c_y$ & rotIoU & aabbIoU \\
\midrule
(a) & $+1$ & $-4$ & $-18$ & $+2$ & 8.6 & 0.82 & $-27$ & $-23$ & 0.53 & 0.66 \\
(b) & $-6$ & $+6$ & $-7$ & $-2$ & 8.2 & 0.79 & $+2$ & $+6$ & 0.60 & 0.87 \\
(c) & $-17$ & $+83$ & $+5$ & $-5$ & 8.2 & 0.00 & $-387$ & $+110$ & 0.00 & 0.00 \\
\bottomrule
\end{tabularx}
\caption{Three %
 illustrative examples from the AVVG test split, rendered from the OBB-PBD seed-1 checkpoint
(Table~\ref{tab:seed-arms}). Each panel shows the full scene (\textbf{top}), in which the yellow box marks the enlarged region, %
 and an enlargement (\textbf{bottom}). Green solid: ground truth. Magenta dashed: OBB-PBD prediction. Cyan dotted: the \texttt{LocateAnything} baseline's axis-aligned prediction, decoded through the same upstream \texttt{generate()}
\textls[15]{path used for the Table~\ref{tab:main-results} baseline row. All outlines are drawn over a black border for
visibility. The~table below the panels gives each prediction's deviation \mbox{from the}} 
}

\label{fig:qualitative}
\end{figure}\vspace{-6pt}
\captionof*{figure}{ground  truth (prediction minus ground truth, in~pixels of  the original $4000\times2250$ image; $|\Delta\theta|$ is the periodic angle error of
Equation~\eqref{eq:angle-mae}); for the axis-aligned baseline, only the center offset and IoUs are meaningful. (\textbf{a}) ``the bottom-most
car'' (OBB-PBD rotIoU 0.82, angle error $8.6^{\circ}$; baseline rotIoU 0.53, aabbIoU 0.66): both models
localize the same car, but~the baseline's axis-aligned box cannot represent its tilt. (\textbf{b})
``the blue bmw mid-size car in the top-middle'' (OBB-PBD rotIoU 0.79, angle error $8.2^{\circ}$; baseline rotIoU
0.60, aabbIoU 0.87): the same pattern in a different scene. (\textbf{c}) ``the brown skoda suv car in the
bottom-middle'' (rotIoU 0.00 for both models): both models place a well-formed box (OBB-PBD angle error
$8.2^{\circ}$) on a different, neighboring car rather than the one referred to. These three
records were selected to illustrate these contrasting behaviors, not sampled at random.}}

\begin{table}[H]
\caption{Main %
 results on the complete AVVG test split (6139 records, 120 images). Every row is evaluated once
on data withheld from checkpoint selection (Section~\ref{sec:evaluation-protocol}). OBB-PBD: mean $\pm$ standard
deviation over three training seeds.}
\label{tab:main-results}
\begin{adjustwidth}{-\extralength}{0cm}
\footnotesize\setlength{\tabcolsep}{2pt}
\begin{tabularx}{\fulllength}{lXcccccc}
\toprule
\textbf{Method} & \textbf{Inference Path} & \textbf{aabbIoU} & \textbf{rotIoU} & \textbf{Acc@0.50} & \textbf{Acc@0.75} & \textbf{Angle MAE (}\boldmath{$^{\circ}$}\textbf{)} & \textbf{Decoded} \\
\midrule
GT HBB envelope (ref.) & GT-derived           & 1.0000          & 0.6037          & 78.6\%          & 16.7\%          & 32.33          &--             \\
LocateAnything (zero-shot) & Upstream \texttt{generate()} & 0.1512   & 0.1253          & 15.1\%          & 3.3\%           &--            & 99.9\%         \\
OBB-PBD (ours)              & Direct PBD             & \textbf{0.334 %
} $\pm$ 0.012 & \textbf{0.303} $\pm$ 0.012 & \textbf{37.1} $\pm$ 1.5\% & \textbf{22.0} $\pm$ 0.8\% & \textbf{9.98} $\pm$ 0.34  & 99.9\%         \\
\bottomrule
\end{tabularx}
\end{adjustwidth}
\noindent{\footnotesize{Bold indicates the best value among the evaluated models; the GT HBB envelope row is a reference computed from the ground truth and is excluded.}}
\end{table}

\section{Results}
\label{sec:results}
\unskip

\subsection{Main Quantitative~Results}
\label{sec:main-results}

We evaluate OBB-PBD on the AVVG test split using the direct PBD decoder. Table~\ref{tab:main-results}
reports the zero-shot baseline, a~GT-derived HBB envelope reference, and~OBB-PBD on the complete AVVG
test split (6139 records, 120 images)---every number in this table comes from locked checkpoints evaluated
once on data that played no role in selecting them. OBB-PBD is reported as the mean and sample standard deviation
over three independently trained seeds (Section~\ref{sec:training-configuration}); per-seed values with
image-cluster confidence intervals are in Table~\ref{tab:seed-arms} (Appendix~\ref{app:seeds}). The~reference row
is a deterministic statistic---the axis-aligned envelope of each target polygon, scored against the oriented
ground truth---not a trained model; Section~\ref{sec:hbb-control} introduces a separate, \emph{trained} HBB-envelope control that predicts this
envelope instead of computing it directly. An~axis-aligned box can still exceed 0.75 rotIoU against a near-square
or near-axis-aligned target, which is why the reference's Acc@0.75 is 16.7\% rather than 0\%.

Relative to the zero-shot baseline, OBB-PBD improves rotIoU by $0.178 \pm 0.012$ (from 0.1253 to 0.303, about
142\% relative, computed as the ratio of the three-seed means) and aabbIoU by $0.183 \pm 0.012$; Acc@0.75 rises
from 3.3\% to 22.0\%. An~image-cluster bootstrap over the 120 test images (resampling images with replacement and
pooling every record belonging to each resampled image, the~correct unit of resampling given within-image
correlation) gives the baseline a rotIoU 95\% confidence interval (CI) of $[0.1073, 0.1460]$ and each OBB-PBD seed
an interval lying entirely above it (Table~\ref{tab:seed-arms}). The~paired per-record improvement over the
baseline, bootstrapped on the same resampled images, excludes zero by a wide margin in every seed: the smallest
lower bounds are $+0.151$ rotIoU and $+0.155$ aabbIoU (Table~\ref{tab:seed-steps}).

This is a cross-path comparison, not a matched ablation: the baseline row runs the upstream \texttt{generate()}
path while OBB-PBD runs this study's direct decoder. Section~\ref{sec:hbb-control} isolates the
orientation-learning effect specifically using fine-tuned controls that share OBB-PBD's data, decoder, and~training~budget.

\subsection{HBB-Envelope Control: Isolating the Format-Switch Effect from the Orientation-Learning~Effect}
\label{sec:hbb-control}

The improvement in Table~\ref{tab:main-results} could in principle come from the six-slot block format itself
rather than from learning to predict orientation. To~separate these two effects, we trained a control arm with a
recipe identical to the reported model---the same block size, six-slot collator, loss weights, and~LoRA targets---except that the training target is the axis-aligned envelope of the ground-truth polygon, so its orientation
is structurally fixed to $0^{\circ}$/$90^{\circ}$. Ground truth for scoring remains the original oriented box in
both arms, so this control isolates the format-switch effect from the orientation-learning effect without changing
the evaluation. Because~a tilted target's envelope is larger than its oriented box, the~orientation-learning step
necessarily compares learning the oriented box with learning its larger envelope; part of its gain is the tighter
extent that representing orientation permits, which cannot be separated from orientation itself. A~third arm,
HBB-PBD, separates the two-slot expansion from the effect of adapting to AVVG itself: it keeps the original
upstream four-slot corner block
(\texttt{<box> x1 y1 x2 y2 </box>}, \texttt{block\_size = 6}) rather than the six-slot block, fine-tuned on the
same subset with the same axis-aligned envelope target as the HBB-envelope control. All three arms are trained
with the same three seeds and use the same validation-disjoint checkpoint-selection protocol
(Section~\ref{sec:evaluation-protocol}); the selected epoch is 5 in eight of the nine trained runs and 4 for
OBB-PBD seed 3, whose validation rotIoU plateaued between epochs 4 and 5. HBB-PBD's evaluation does not exclude
corner-touching targets: the four-slot format cannot represent
axis-aligned envelopes that touch a corner of the resized image (about 4\% of this test split), and~those cases
are scored as ordinary decode failures (IoU~0) rather than excluded, so this arm is evaluated on the same test records as
every row of Table~\ref{tab:main-results}. Its own training set is nonetheless about 10.3\% smaller
than the HBB-envelope control's and the main model's (20{,}182 vs.\ 22{,}495 AVVG training records), because~the
same corner-frame constraint that is tolerated as a decode failure at test time was applied as an exclusion
filter at train time; this arm's numbers should be read with that training-data asymmetry in~mind.

Paired image-cluster bootstraps (the same resampling procedure as Section~\ref{sec:main-results}, applied to each
step's per-record delta, separately for each seed; Table~\ref{tab:seed-steps}) give the same picture in every
seed. The~adaptation step's rotIoU and aabbIoU intervals exclude zero in all three seeds. The~format-switch
step's rotIoU interval includes zero in all three seeds ($-0.005$ to $+0.000$): expanding the block from four
slots to six without learning orientation produces no rotIoU improvement distinguishable from zero, despite the
two arms' training-data-size asymmetry noted above. The~orientation-learning step's rotIoU interval excludes zero
in all three seeds ($+0.046$ to $+0.060$; smallest lower bound $+0.036$), an~18.7--23.7\% relative increase over
the matched control's own rotIoU. The~gain does not depend on the six-slot control alone: OBB-PBD also exceeds the
four-slot HBB-PBD arm in every seed ($+0.041$ to $+0.060$ rotIoU, every interval excluding zero).

aabbIoU moves much less than rotIoU across the three fine-tuned arms. The~HBB-envelope control is $0.012 \pm 0.003$
below HBB-PBD (interval excluding zero in \mbox{seeds 2 and 3}), and~OBB-PBD is $0.010 \pm 0.010$ below the HBB-envelope
control ($-0.002$ to $-0.021$). Learning orientation therefore improves rotated alignment without improving
axis-aligned localization; if anything, axis-aligned agreement is slightly~lower.

Most of the baseline-to-final gain thus comes from the first step, the~joint effect of domain adaptation on
AVVG and the change of decoding path, which a four-slot model shows too (Table~\ref{tab:decomposition} $^{\dagger}$);
the present comparisons do not separate these two parts. The~six-slot block format adds no distinguishable gain,
and orientation learning adds a smaller further gain (Table~\ref{tab:decomposition}). The~zero-shot
baseline comparison in Table~\ref{tab:main-results} remains this paper's primary result; the decomposition here
is evidence for the narrower claim that learning the oriented box rather than its axis-aligned envelope---not
merely having more coordinate slots---is responsible for part of that improvement, at~a small cost in
axis-aligned agreement. The~HBB-PBD arm's training-data-size confound noted above
still applies to this~comparison.

\begin{table}[H]
\caption{Decomposition of the reported improvement into an adaptation step (domain adaptation combined with a
change of decoding path), a~format-switch step, and~an orientation-learning step, on~the complete AVVG test split (6139 records, 120 images). Each step is the paired
per-record difference between consecutive arms, given as mean $\pm$ standard deviation over three training seeds;
per-seed values with image-cluster confidence intervals are in Table~\ref{tab:seed-steps}. The~orientation-learning step includes the tighter extent an oriented box allows (Section~\ref{sec:hbb-control}).}
\label{tab:decomposition}

\begin{adjustwidth}{-\extralength}{0cm}

\setlength{\cellWidtha}{ \fulllength/3-2\tabcolsep--1.40in}
\setlength{\cellWidthb}{ \fulllength/3-2\tabcolsep-.70in}
\setlength{\cellWidthc}{ \fulllength/3-2\tabcolsep-.70in}
\scalebox{1}[1]{\begin{tabularx}{ \fulllength}{>{\raggedright\arraybackslash}m{\cellWidtha}>{\centering\arraybackslash}m{\cellWidthb}>{\centering\arraybackslash}m{\cellWidthc}}
\toprule
\textbf{Step} & \boldmath{$\Delta$}\textbf{aabbIoU} & \boldmath{$\Delta$}\textbf{rotIoU} \\
\midrule
Baseline $\rightarrow$ HBB-PBD, 4-slot (adaptation + decoding path) $^{\dagger}$  & $+0.205 \pm 0.001$ & $+0.125 \pm 0.002$ \\
HBB-PBD $\rightarrow$ HBB-envelope, 6-slot (format-switch effect)        & $-0.012 \pm 0.003$ & $-0.002 \pm 0.003$ \\
HBB-envelope $\rightarrow$ ours (orientation-learning effect)            & $-0.010 \pm 0.010$ & \textbf{$+0.055 \pm 0.008$} \\
\midrule
Baseline $\rightarrow$ ours (total, Table~\ref{tab:main-results})       & $+0.183 \pm 0.012$ & $+0.178 \pm 0.012$ \\
\bottomrule
\end{tabularx}}
\end{adjustwidth}

\noindent\small{$^{\dagger}$ Joint effect of domain adaptation and a decoding-path change: the baseline runs the
upstream \texttt{generate()} path, while every trained arm uses this study's direct decoder, so this row does not
isolate fine-tuning alone.}
\end{table}

\subsection{Image-Dependence~Diagnostic}
\label{sec:image-dependence}

To verify that the trained model uses the image rather than relying on prompt statistics or a mean-box prior,
we repeated the evaluation after shuffling images and prompts independently, on~the complete test split (6139
records, 120 images) using the same image-cluster bootstrap as Section~\ref{sec:main-results}. The~diagnostic
was repeated for each of the three OBB-PBD seeds; the matched condition reproduces each seed's main result, and~in
every seed, the matched and image-shuffled confidence intervals are disjoint by a wide~margin.

In each seed, the matched condition is 46--51 times the image-shuffled rotIoU (49.96, 45.79 and 50.96 from
unrounded values), and~the angle error increases by about $29.5^{\circ}$ after image shuffling. This is evidence
of image dependence under the validated~pairing.

The image-shuffled condition still retains a small rotIoU (0.0057--0.0067 across seeds), so we examined whether it reflects prior
knowledge of the prompts or of typical target locations. Because~the derangement is over records and each test
image carries about 51 records, 1.1\% of shuffled records received another record's image that happens to be their
own; these records score 0.22--0.29 rotIoU and account for 44--50\% of the residual across the three seeds.
Excluding them, the~residual is 0.0032--0.0034 (seed 1: 95\% CI $[0.0027, 0.0041]$), which matches a model-free
location prior---predicting for every record the mean target box of the test records from all other images---at
0.0038 ($[0.0024, 0.0052]$). The~remaining residual is
therefore at the level of a generic location-and-size prior, not evidence of grounding from the text~alone.

\subsection{Efficiency and Limitations of the~Result}
\label{sec:efficiency}

Latency is measured on a single GPU (NVIDIA RTX 6000 Ada, 48\,GB) at batch size one over 60 test samples. The~six-slot rows use the OBB-PBD seed-1 checkpoint (Table~\ref{tab:seed-arms}); the four-slot row uses the base
weights in the upstream four-slot layout, since latency depends on the block configuration rather than on the
fine-tuned weight values. Expanding the parallel coordinate block from four HBB slots to six OBB slots increases
latency from 306.87~ms to 329.82~ms (median 306.58~ms and 329.77~ms, 95th
percentile 309.45~ms and 332.57~ms), a~$1.075\times$ ratio---a modest slot-expansion~overhead.

To measure what parallel decoding saves, we also decode the same six geometry tokens autoregressively. This
baseline runs one full forward pass over the image, prompt, and \texttt{<box>} token, then emits the six geometry
tokens and the closing \texttt{</box>} one at a time, each step running only the language model on the previous
greedy token with the cached keys and values. The~vision encoder therefore runs once, as~in standard cached
generation, and~both paths produce the same number of output tokens. Autoregressive decoding takes 568.65~ms
(median 568.85~ms, 95th percentile 571.13~ms), so one parallel pass is $1.72\times$ faster for the same six
coordinates. The~gain is smaller than the token count alone would suggest because the shared prefix, including
the vision encoder, dominates the cost of both~paths.

The main accuracy limitation is referring failures in dense aerial scenes. Qualitative overlays
(Figure~\ref{fig:qualitative}) and the per-sample analysis show cases where the predicted geometry is close but
the prediction is placed on a neighboring object. The~AVVG prompts also contain subjective attributes such as ``the cheapest car,'' which
cannot always be resolved from box geometry alone. Further limitations are discussed in
Section~\ref{sec:limitations}.

\section{Discussion}
\label{sec:discussion}
\unskip

\subsection{Decomposition of Format and Orientation~Effects}
\label{sec:discussion-decomposition}

The HBB-envelope control tests whether expanding the coordinate block from four to six slots could explain the
improvement without learning orientation. The~format-switch comparison (Table~\ref{tab:decomposition}) provides no
evidence that block expansion alone improves rotIoU: its 95\% CI includes zero in every seed, although~the
HBB-PBD arm was trained on fewer samples. By~contrast, learning orientation, compared with an otherwise matched
control, adds $0.055 \pm 0.008$ rotIoU over three seeds, with~each seed's image-cluster CI excluding zero, an~18.7--23.7\% relative increase over that control's own rotIoU. Because~the control learns the axis-aligned
envelope, this gain covers both the angle and the tighter extent an oriented box allows. Most of the
baseline-to-final gain arises earlier, in~the joint step of domain adaptation and decoding-path change, which a
four-slot model shows too; that comparison does not isolate fine-tuning alone
(Table~\ref{tab:decomposition} $^{\dagger}$).

The aabbIoU results separate localization from orientation. The~adaptation step raises aabbIoU by about 0.2 in
every arm, whereas the differences among the three fine-tuned arms are an order of magnitude smaller (0.002--0.036),
with OBB-PBD slightly below the axis-aligned controls. The~adaptation step therefore accounts for most of the
axis-aligned localization gain, while the orientation slots improve rotated alignment specifically. Reporting aabbIoU preserves
comparability with the horizontal-box baseline, but~the benefit of orientation learning appears only in rotIoU;
in aabbIoU, it comes with a small cost (up to 0.021 below the HBB-envelope control, with~the interval excluding
zero in seed 3).

\subsection{Limitations}
\label{sec:limitations}

First, this study is limited to the validated AVVG/UAV setting. Its results constitute a controlled benchmark
under that protocol (Section~\ref{sec:evaluation-protocol}), not evidence of generalization across datasets,
sensors, object scales, or grounding annotation formats, nor of the optimality of OBBs or PBD beyond that
setting.

Second, referring failures in dense aerial scenes remain the main observed error mode (Section~\ref{sec:efficiency}).

Third, this study does not compare head-to-head against joint-regression architectures for oriented-box
grounding that use localization tokens and a multilayer perceptron (MLP) head, such as SATGround \citep{toker2025satground}
(Section~\ref{sec:rw-parallel-decoding}). Such a comparison would require reimplementing a different architecture
family under this study's backbone, data, and evaluation protocol, which is outside the scope of characterizing
the specific PBD block extension studied here (Section~\ref{sec:rw-positioning}). The claim in this paper is
accordingly limited to the cost and accuracy of extending LocateAnything's block-aligned decoding interface to
orientation; it is not a claim of superiority over joint OBB regression.

Fourth, Angle MAE (Table~\ref{tab:main-results}, Table~\ref{tab:image-dependence}) averages over all successfully
decoded samples, including those with $r_{\text{norm}}<1.05$, for which the directional loss assigns zero
gradient (Equation~\eqref{eq:loss-direction}). As Section~\ref{sec:training-configuration} notes, on AVVG's
non-square images this gate selects boxes near the image's aspect ratio rather than physically near-square boxes.
Because these samples can still contribute large periodic-distance errors, Angle MAE should be read as an
aggregate over the full test population, not as an orientation-quality diagnostic restricted to the subset
optimized by the directional loss.

\begin{table}[H]
\caption{{Image}-dependence diagnostic on the complete AVVG test split: mean $\pm$ standard deviation over the three
OBB-PBD seeds, with~the range of the per-seed image-cluster 95\% CIs.}
\label{tab:image-dependence}
\begin{tabularx}{\textwidth}{>{\raggedright\arraybackslash}Xcccc}
\toprule
\textbf{Condition} & \textbf{rotIoU} & \textbf{Per-Seed 95\% CIs} & \textbf{Acc@0.75} & \textbf{Angle MAE (}\boldmath{$^{\circ}$}\textbf{)} \\
\midrule
Matched image and prompt & \textbf{0.303 %
} $\pm$ 0.012 & $0.269$--$0.334$ & \textbf{22.0} $\pm$ 0.8\% & \textbf{9.98} $\pm$ 0.34 \\
Image shuffled            & 0.006 $\pm$ 0.001          & $0.004$--$0.008$ & 0.2 $\pm$ 0.0\%           & 39.46 $\pm$ 0.19          \\
Prompt shuffled           & 0.044 $\pm$ 0.002          & $0.033$--$0.056$ & 3.1 $\pm$ 0.3\%           & 19.18 $\pm$ 0.28          \\
\bottomrule
\end{tabularx}
\noindent{\footnotesize{Bold marks the highest rotIoU and Acc@0.75 and the lowest Angle MAE (the matched condition).}}
\end{table}

\section{Conclusions}
\label{sec:conclusions}

This paper extends LocateAnything's Parallel Box Decoding interface from a four-coordinate horizontal box to a
six-slot oriented box and reports what the extension costs and gains under a diagnostic evaluation protocol.
Diagnostic probes exposed a backbone defect that let the training loss decrease while the coordinate slots could
not see the image, and~verified its correction; shuffle controls and visibility/leakage probes check the reported
decoding path throughout. Under~this protocol, and~across three training seeds, the~extension improves rotIoU over
the zero-shot baseline on the complete AVVG test split, with~checkpoint selection kept disjoint from that split
(Section~\ref{sec:evaluation-protocol}). Matched control arms isolate how much of this gain comes from learning the
oriented box rather than from the larger coordinate block alone (Section~\ref{sec:hbb-control}).

Two methodological directions follow directly from what this study leaves open. First, future work should
validate the loss weights in Table~\ref{tab:training-settings} through multi-seed, equal-budget ablations of the
reported five-epoch configuration. Second, a~matched-protocol comparison with localization-token/MLP
joint-regression architectures such as SATGround \citep{toker2025satground} (Section~\ref{sec:limitations}) would
quantify how coordinate-token decoding and dedicated regression heads differ in accuracy, decoded percentage,
and inference cost. Beyond~these, extending the validated protocol to remote-sensing settings outside
the AVVG/UAV data studied here---different sensors, object scales, or~grounding annotation formats---remains untested and
is not claimed~here.

\vspace{6pt}
\authorcontributions{%
Conceptualization, methodology, software, validation, formal analysis,
investigation, data curation, writing---original draft preparation, and~visualization, C.W.S.; writing---review and editing and
supervision, Y.J.K. All authors have read and agreed to the
published version of the manuscript.}

\funding{%
This work was partly supported by the National Research Foundation of Korea (NRF) grants funded by the Korea government (MSIT) (No. RS-2025-00559165), and partly by Institute of Information \& communications Technology Planning \& Evaluation (IITP) grant funded by the Korea government (MSIT) (No. RS-2022-00155857, Artificial Intelligence Convergence Innovation Human Resources Development (Chungnam National University)). %
}

\institutionalreview{%
Not applicable. %
}

\dataavailability{%
Publicly available datasets were analyzed in this study. The~AVVG subset of RefGeo used for training and
evaluation is available at %
 \url{https://huggingface.co/datasets/erenzhou/refGeo} under the S-Lab License 1.0. The~LocateAnything-3B backbone extended in this work is available {at}
\url{https://huggingface.co/nvidia/LocateAnything-3B} under the NVIDIA License.
Code implementing the OBB-PBD extension, training configuration, and~evaluation protocol described in this paper,
together with trained model checkpoints under the same NVIDIA License and Qwen RESEARCH LICENSE AGREEMENT
conditions that apply to the LocateAnything-3B backbone they extend, will be available at \url{https://github.com/ce-dric/OBB-PBD} and
\url{https://huggingface.co/ce-dric/OBB-PBD}.
A unified source diff of the five fixed upstream defects (Section~\ref{sec:upstream-defects}) against the pinned
pristine upstream revision, and~a post-run package inventory captured from the retained training container (Ubuntu 22.04.5, Python 3.11.10, PyTorch 2.5.1+cu121, Transformers 4.57.1), will be included in that same   code release.}

\acknowledgments{%
During the preparation of this study, the~authors used Claude Code (Anthropic; Claude Sonnet 5 and Claude Opus 5.5) %
 and Codex (OpenAI; GPT-6-Sol) to assist with
implementation, execution of author-approved experiments, and~manuscript drafting, and~Codex and Antigravity (Google; Gemini 3.8 Flash)
for independent manuscript cross-checks; model versions varied over the course of the study
(2026). The~authors reviewed, tested, and~edited all AI-assisted outputs and take full responsibility for the
content of this~publication.}

\conflictsofinterest{%
The authors declare no conflicts of~interest.}

\appendixtitles{yes}
\appendixstart
\appendix
\section[\appendixname~\thesection]{Upstream Implementation Defects}
\label{app:defects}

Table~\ref{tab:defects} lists the backbone corrections referred to in Section~\ref{sec:upstream-defects}. U1 only
affects configurations with more than four coordinate slots; U2--U5 affect any fine-tuning of this~backbone.

\begin{table}[H]
\caption{Corrections to the pretrained \texttt{LocateAnything-3B} code made during this~study.}
\label{tab:defects}
\begin{adjustwidth}{-\extralength}{0cm}
\footnotesize
\begin{tabularx}{\fulllength}{cXX}
\toprule
\textbf{ID} & \textbf{Symptom} & \textbf{Cause and Correction} \\
\midrule
U1 & Eight-token block cannot be constructed. &
Block size capped by a hard-coded \texttt{min(4, block\_size)}; now read from the~configuration. \\
\midrule
U2 & Crash in training mode. &
Forward pass returned a tuple instead of \texttt{CausalLMOutputWithPast}; return type~corrected. \\
\midrule
U3 & Backward pass fails (in-place write into a leaf view). &
Embeddings reshaped for image-token injection remained a leaf view; \texttt{.clone()} added. \\
\midrule
U4 & Evaluation loss crashes when \texttt{input\_ids=None}. &
Mask selection keyed only on \texttt{self.training}; \texttt{input\_ids=None} calls now use the training~mask. \\
\midrule
U5 & No crash; loss decreases, but~coordinate slots see no image or prompt tokens (on one batch, the~first
coordinate slot attended to 7 of 894 keys; 894 of 894 after the fix). &
Causal-prefix length derived from position-ID drops that training never produces ($-1$ for every sample); now
recovered from the mask-token region, as~at~inference. \\
\bottomrule
\end{tabularx}
\end{adjustwidth}
\end{table}
\unskip

\section[\appendixname~\thesection]{Per-Seed Results}
\label{app:seeds}

Tables~\ref{tab:seed-arms} and \ref{tab:seed-steps} give the per-seed values behind the means in
Tables~\ref{tab:main-results}--\ref{tab:decomposition}. Seeds 1--3 use training seeds 20260812, 20260813, and
20260814 with distinct data-order seeds; the image-level split is identical across seeds. Intervals are 95\%
image-cluster bootstrap CIs (120 test images, 10{,}000 draws).

\begin{table}[H]
\caption{Per%
-seed results of the three trained arms on the complete AVVG test split (6139 records, 120 images).
Zero-shot baseline for reference: rotIoU 0.1253 $[0.1073, 0.1460]$, aabbIoU 0.1512.}
\label{tab:seed-arms}
\begin{adjustwidth}{-\extralength}{0cm}
\footnotesize
\begin{tabularx}{\fulllength}{lccXXcc}
\toprule
\textbf{Arm} & \textbf{Seed} & \textbf{Epoch} & \textbf{rotIoU [95\% CI]} & \textbf{aabbIoU [95\% CI]} & \textbf{Acc@0.75} & \textbf{Angle MAE ($^{\circ}$)} \\
\midrule
OBB-PBD & 1 & 5 & 0.3124 $[0.2918, 0.3342]$ & 0.3441 $[0.3234, 0.3659]$ & 22.8\% & 9.59 \\
 & 2 & 5 & 0.3062 $[0.2867, 0.3273]$ & 0.3378 $[0.3184, 0.3585]$ & 22.1\% & 10.16 \\
 & 3 & 4 & 0.2898 $[0.2693, 0.3117]$ & 0.3207 $[0.3000, 0.3426]$ & 21.2\% & 10.20 \\
\midrule
HBB-envelope & 1 & 5 & 0.2527 $[0.2364, 0.2702]$ & 0.3465 $[0.3261, 0.3682]$ & 3.1\% & 22.01 \\
 & 2 & 5 & 0.2476 $[0.2321, 0.2637]$ & 0.3432 $[0.3229, 0.3644]$ & 2.9\% & 22.31 \\
 & 3 & 5 & 0.2442 $[0.2275, 0.2624]$ & 0.3420 $[0.3208, 0.3642]$ & 2.3\% & 22.09 \\
\midrule
HBB-PBD & 1 & 5 & 0.2523 $[0.2358, 0.2696]$ & 0.3552 $[0.3337, 0.3780]$ & 4.0\% & 30.57 \\
 & 2 & 5 & 0.2484 $[0.2322, 0.2655]$ & 0.3572 $[0.3345, 0.3812]$ & 3.9\% & 30.57 \\
 & 3 & 5 & 0.2493 $[0.2335, 0.2660]$ & 0.3564 $[0.3349, 0.3786]$ & 4.1\% & 29.89 \\
\bottomrule
\end{tabularx}
\end{adjustwidth}
\end{table}
\unskip

\begin{table}[H]
\caption{Per%
-seed paired differences between arms (later arm minus earlier arm, per record), with~95\%
image-cluster CIs. The~last block compares OBB-PBD directly with the four-slot HBB- PBD arm.}
\label{tab:seed-steps}
\begin{adjustwidth}{-\extralength}{0cm}
\footnotesize

\setlength{\cellWidtha}{ \fulllength/4-2\tabcolsep--.90in}
\setlength{\cellWidthb}{ \fulllength/4-2\tabcolsep-.90in}
\setlength{\cellWidthc}{ \fulllength/4-2\tabcolsep-0in}
\setlength{\cellWidthd}{ \fulllength/4-2\tabcolsep-0in}
\scalebox{1}[1]{\begin{tabularx}{ \fulllength}{>{\raggedright\arraybackslash}m{\cellWidtha}>{\centering\arraybackslash}m{\cellWidthb}>{\raggedright\arraybackslash}m{\cellWidthc}>{\raggedright\arraybackslash}m{\cellWidthd}}

\toprule
\textbf{Step} & \textbf{Seed} & \boldmath{$\Delta$}\textbf{rotIoU [95\% CI]} & \boldmath{$\Delta$}\textbf{aabbIoU [95\% CI]} \\
\midrule
Baseline $\rightarrow$ HBB-PBD (adaptation + decoding path)$^{\dagger}$ & 1 & $+0.1271$ $[+0.1128, +0.1410]$ & $+0.2040$ $[+0.1893, +0.2187]$ \\
 & 2 & $+0.1231$ $[+0.1098, +0.1361]$ & $+0.2060$ $[+0.1912, +0.2206]$ \\
 & 3 & $+0.1241$ $[+0.1100, +0.1376]$ & $+0.2052$ $[+0.1903, +0.2198]$ \\
\midrule
HBB-PBD $\rightarrow$ HBB-envelope (format switch) & 1 & $+0.0004$ $[-0.0076, +0.0085]$ & $-0.0087$ $[-0.0188, +0.0017]$ \\
 & 2 & $-0.0009$ $[-0.0095, +0.0079]$ & $-0.0140$ $[-0.0263, -0.0016]$ \\
 & 3 & $-0.0051$ $[-0.0139, +0.0030]$ & $-0.0143$ $[-0.0260, -0.0034]$ \\
\midrule
HBB-envelope $\rightarrow$ OBB-PBD (orientation learning) & 1 & $+0.0597$ $[+0.0495, +0.0702]$ & $-0.0024$ $[-0.0112, +0.0066]$ \\
 & 2 & $+0.0587$ $[+0.0488, +0.0689]$ & $-0.0054$ $[-0.0135, +0.0025]$ \\
 & 3 & $+0.0456$ $[+0.0359, +0.0560]$ & $-0.0214$ $[-0.0315, -0.0106]$ \\
\midrule
Baseline $\rightarrow$ OBB-PBD (total) & 1 & $+0.1872$ $[+0.1742, +0.2004]$ & $+0.1930$ $[+0.1797, +0.2062]$ \\
 & 2 & $+0.1810$ $[+0.1681, +0.1937]$ & $+0.1866$ $[+0.1729, +0.1998]$ \\
 & 3 & $+0.1645$ $[+0.1512, +0.1780]$ & $+0.1695$ $[+0.1554, +0.1834]$ \\
\midrule
HBB-PBD $\rightarrow$ OBB-PBD (check) & 1 & $+0.0601$ $[+0.0489, +0.0719]$ & $-0.0111$ $[-0.0203, -0.0015]$ \\
 & 2 & $+0.0578$ $[+0.0458, +0.0696]$ & $-0.0194$ $[-0.0314, -0.0075]$ \\
 & 3 & $+0.0405$ $[+0.0277, +0.0540]$ & $-0.0357$ $[-0.0465, -0.0245]$ \\
\bottomrule
\end{tabularx}}
\end{adjustwidth}
\noindent\small{$^{\dagger}$ Joint effect of domain adaptation and a decoding-path change; does not isolate fine-tuning
alone (Table~\ref{tab:decomposition}).}
\end{table}

\begin{adjustwidth}{-\extralength}{0cm}

\reftitle{References}

\PublishersNote{}
\end{adjustwidth}
\end{document}